\documentclass[12pt]{article}

\usepackage[utf8]{inputenc}
\usepackage[T1]{fontenc}
\usepackage{lmodern}        
\usepackage{setspace}       
\usepackage{geometry}
\usepackage{natbib}         
\usepackage{url}            
\usepackage{graphicx}       
\usepackage{tikz}           
\usetikzlibrary{shapes,arrows,positioning}
\usepackage{verbatim}       
\usepackage{xcolor}         
\usepackage{listings}       
\usepackage{hyperref}       

\title{\textbf{Generative AI in Education:\\
From Foundational Insights to the Socratic Playground for Learning}}
\author{Xiangen Hu, Sheng Xu, Richard Tong, \& Art Graesser}
\date{\today}

\begin{document}

\maketitle
\doublespacing

\begin{abstract}
This paper explores the synergy between human cognition and Large Language Models (LLMs), highlighting how generative AI can drive personalized learning at scale. We discuss parallels between LLMs and human cognition, emphasizing both the promise and new perspectives on integrating AI systems into education. After examining challenges in aligning technology with pedagogy, we review AutoTutor—one of the earliest Intelligent Tutoring Systems (ITS)—and detail its successes, limitations, and unfulfilled aspirations. We then introduce the Socratic Playground, a next-generation ITS that uses advanced transformer-based models to overcome AutoTutor’s constraints and provide personalized, adaptive tutoring. To illustrate its evolving capabilities, we present a JSON-based tutoring prompt that systematically guides learner reflection while tracking misconceptions. Throughout, we underscore the importance of placing pedagogy at the forefront, ensuring that technology’s power is harnessed to enhance teaching and learning rather than overshadow it.
\end{abstract}

\section{Introduction and Background}
\label{sec:intro}
Recent advancements in artificial intelligence (AI), particularly Large Language Models (LLMs) \citep{zhao2023survey}, have sparked renewed interest in understanding both the potential and the changing viewpoints on these technologies in education . As researchers draw parallels between LLMs and human cognition, new opportunities emerge for enhancing personalized learning, but persistent questions remain about aligning such innovations with sound pedagogical practice \citep{memarian2023chatgpt}. In this section, we examine the cognitive foundations of LLMs, consider how generative AI can outpace human performance, and highlight why technology integration often stalls unless carefully woven into educational frameworks.

\subsection{Human Cognition and Large Language Models (LLMs)}
The study of human cognition has long been a cornerstone of educational research, focusing on processes such as reasoning, problem-solving, and language comprehension. Recent developments in AI, particularly LLMs, reveal striking parallels with human cognitive functions. Like humans, LLMs rely on contextual understanding and pattern recognition to process and generate coherent responses. The Transformer architecture, a foundation for many state-of-the-art models, uses mechanisms analogous to attention and memory in the human brain \citep{Vaswani2017}. These parallels suggest a bidirectional opportunity: insights into LLMs can enhance our understanding of cognition, while principles of human learning can guide the development of AI technologies. Recent advancements in neuro-symbolic cognitive architectures, such as NEOLAF (Never-Ending Open Learning Adaptive Framework), exemplify this bidirectional synergy. By integrating symbolic reasoning with neural learning, NEOLAF mirrors human cognition’s adaptability while enabling LLMs to function as self-improving agents \citep{tong2023neolaf}. This approach highlights how AI technologies can transcend traditional models of pattern recognition, offering a deeper alignment with human cognitive processes. By leveraging this synergy, educational frameworks can better emulate human-like adaptability, fostering more personalized and effective learning experiences. 

\subsection{Generative AI’s Potential Beyond Human Cognitive Limits}
Generative AI technologies, exemplified by OpenAI's o3 model, are redefining cognitive boundaries. The o3 model, introduced in December 2024, excels in tasks requiring deep reasoning, such as mathematics, scientific problem-solving, and competitive programming, achieving human-equivalent or superior results. For instance, it scored 96.7\% on the 2024 American Invitational Mathematics Exam (missing just one question) and demonstrated expertise comparable to doctoral candidates in physics, chemistry, and biology. This unprecedented capability highlights the potential of generative AI to exceed human cognitive performance, not only at the individual level but also by synthesizing collective knowledge. In education, these advancements promise revolutionary applications, from personalized tutoring systems to automated content creation \citep{maity2024generative}, enabling scalable solutions to global learning challenges. However, these developments also call for a critical evaluation of their integration into pedagogical systems, ensuring that human development remains central to technological progress.

\subsection{Challenges in Aligning Technology and Pedagogy}
Despite the rapid evolution of educational technologies, their success depends fundamentally on how effectively they align with pedagogical principles. Education systems traditionally prioritize human intellectual growth, with teaching methods designed to cultivate critical thinking, creativity, and problem-solving skills \citep{NationalResearchCouncil2000}. However, technological innovations often fail to directly enhance learning unless thoughtfully integrated. \citet{KoehlerMishra2009} highlight the necessity of blending technological tools with pedagogical strategies to achieve meaningful educational outcomes. For instance, adaptive learning platforms show great promise but often lack the nuanced understanding of learners’ needs that skilled educators bring. Bridging this gap requires not only advanced tools but also a pedagogy-first approach, emphasizing the educator’s role in mediating technology’s impact on learning.

\subsection{Efficiency of Technology for Educators vs. Learning Outcomes for Students}
While technology has undeniably improved educators’ efficiency, its impact on students’ learning outcomes remains inconsistent. Tools like automated grading systems and content recommendation algorithms can reduce educators' workloads, enabling them to focus on more interactive teaching practices \citep{Perrotta2013}. However, these efficiencies do not inherently translate into improved learning for students. Research shows that student outcomes are highly dependent on active engagement and self-directed learning, which require intentional pedagogical interventions \citep{Chi2009}. Moreover, the assumption that technology alone can replace the complex relational and motivational elements of teaching is fundamentally flawed. To maximize the benefits of educational technologies, design strategies must prioritize student-centered approaches, ensuring that innovations empower learners to actively construct knowledge rather than passively receive information.

\subsection*{Integrating Advanced LLMs: An Evolving Perspective}
\noindent
In sum, the parallels between LLMs and human cognition present a compelling rationale for leveraging conversation-based intelligent tutoring: if humans learn through dialogue, then LLMs—whose linguistic operations approximate key cognitive processes—can effectively enrich that experience. Given that many state-of-the-art models now exceed human capabilities in specialized tasks, cautionary emphasis on their limitations is less vital than harnessing their advanced reasoning to enhance learning. Yet even with access to the most powerful LLMs, it is thoughtful pedagogy—rather than sheer computational sophistication—that ultimately drives genuine educational outcomes. Unless technology is coupled with sound instructional strategies, its impact on learner growth remains superficial. The same principle applies to educators’ broader adoption of digital tools: merely adding more technology does not guarantee transformative teaching. Instead, it is by embedding carefully chosen AI resources within robust pedagogical frameworks that we realize lasting improvements in student engagement and understanding. This perspective will underlie the coming sections and resonate again in our final discussion, underscoring how true innovation in education hinges on the synergy of advanced AI with well-designed teaching approaches.

\section{Generative AI in Education}
\label{sec:genAI}
Generative AI represents a transformative force in education, offering tools and methodologies that redefine how knowledge is created, disseminated, and consumed. Although earlier discourse often focused on the limitations of LLMs, we now have good reason to consider the most advanced models in harnessing their capabilities for instructional gains. This section explores how generative AI can bridge cognitive gaps, augment educators, and promote equitable learning, serving as a powerful complement to sound pedagogy.

\subsection{Bridging Cognitive Gaps}
Generative AI systems, such as large language models, are reshaping personalized education by aligning content delivery with individual learning needs. These models excel at analyzing user inputs to identify gaps in knowledge, misunderstandings, or skill deficits and adapt their outputs accordingly. This dynamic interaction allows learners to receive tailored content and explanations that address their unique cognitive challenges, fostering a deeper understanding of complex topics\citep{cooper2023examining} \citep{baidoo2023education}. Neuro-symbolic systems like NEOLAF further expand this potential by leveraging domain-specific reasoning to refine content delivery. These systems provide adaptive scaffolding by incorporating symbolic knowledge representations alongside neural learning, addressing learner misconceptions with precision and supporting iterative cognitive development \citep{tonghu2024neurosymbolic}. As such, they align closely with principles of human learning while benefiting from the efficiency of generative AI.

For example, AI-powered platforms can scaffold learning experiences by breaking down challenging concepts into manageable steps, a technique rooted in Vygotsky’s scaffolding theory \citep{Vygotsky1978}. By offering real-time support, advanced LLMs help learners build foundational knowledge incrementally, reducing cognitive overload \citep{Shadiev03032020} and promoting long-term retention \citep{Cepeda2006}\citep{educsci13121216}. Furthermore, the adaptive nature of these systems aligns with spaced repetition and active recall methodologies, proven to enhance learning outcomes.

Generative AI is also instrumental in creating diverse representations of knowledge. By generating analogies, rephrased explanations, or multimodal content (text, visuals, and even simulations), these tools cater to different learning styles and preferences, addressing variability in cognitive and cultural backgrounds \citep{Mayer2002}. However, as the final subsection of the previous chapter notes, these systems require robust pedagogy to avoid becoming mere information dispensers. Technology alone does not suffice to drive authentic learning experiences.

\subsection{Augmenting Educators}
The role of educators in the 21st century is increasingly multifaceted, encompassing teaching, mentorship, and the curation of personalized learning experiences. Generative AI serves as a valuable ally in this endeavor, freeing educators from repetitive or administrative tasks and enabling them to focus on high-impact activities that require human judgment and emotional intelligence.

AI systems can also assist educators in designing instructional materials that align with curriculum standards while integrating innovative elements. For example, generative models can produce customized quizzes, case studies, or project ideas tailored to specific learning objectives and student demographics \citep{Luckin2016}. Natural language interfaces play a pivotal role in dialogue systems, enabling intuitive interaction between students, educators, and AI systems. As noted by Tong and Li \citep{TongLee2023}, these interfaces form the backbone of trustworthy AI systems for education, facilitating Human-in-the-Loop processes that enhance assisted learning and targeted student interventions. By engaging learners in dynamic, conversational exchanges, such systems provide real-time insights into student needs, enabling educators to address misconceptions and adapt teaching strategies effectively. These tools further enable educators to experiment with novel pedagogical approaches, such as gamified learning or scenario-based simulations, which are otherwise resource-intensive to develop.

Yet, consistent with the perspective that advanced LLMs thrive when paired with thoughtful human oversight, teachers still need professional development support. They must learn not only how to operate AI tools but also how to scrutinize their outputs. Without careful implementation, even sophisticated LLMs can yield superficial learner engagement or, worse, misdirected feedback \citep{prather2024widening}\citep{zhai2024effects}.

\subsection{Risks of Over-Reliance and the Focus on Pedagogy}
While advanced generative AI has made strides in reliability and performance, there remain fundamental educational and human-centric considerations. Rather than overly magnifying limitations of LLMs themselves, the real pitfall may be an over-reliance on technology in lieu of strong pedagogical frameworks. Education thrives on an iterative process of questioning, reasoning, and understanding—an approach made stronger when thoughtful educators guide learners rather than delegating entirely to machines \citep{KASNECI2023102274}.

Students accustomed to unquestioning acceptance of AI-generated answers may struggle to cultivate independent critical thinking skills. Similarly, educators who rely too heavily on AI tools risk underestimating the importance of fostering inquiry-driven classroom cultures. Efficiency gains in grading or content generation must therefore be balanced by active strategies that encourage learner autonomy, reflection, and peer collaboration.

\subsection{Enhancing Accessibility and Equity}
Generative AI has the potential to bridge significant educational divides, offering unprecedented opportunities for learners in underserved and remote regions \citep{li2023empowering}. Advanced AI-powered platforms can provide high-quality instructional resources that are scalable and cost-effective, accessible through digital devices. For example, tools like Khanmigo—an AI tutor integrated into Khan Academy—exemplify how these innovations can open personalized learning experiences for large numbers of users \citep{Heffernan2020}\citep{shetye2024evaluation}.

Moreover, AI systems can excel at localizing and customizing content to suit diverse linguistic and cultural contexts \citep{athanassopoulos2023use}. By generating translations, culturally relevant examples, and adaptive explanations, these tools make education more inclusive for non-native speakers and marginalized communities. This capability addresses long-standing inequities in educational access and quality, particularly in low-income regions where resources are scarce.

However, aligning technology with pedagogy remains essential. Without effective instructional design, even the most powerful LLM-based systems will not guarantee successful learning outcomes. Equitable adoption of AI thus demands both improved digital infrastructure and an unwavering focus on high-quality teaching practices.

\section{Synergy of Technology and Pedagogy}
\label{sec:synergy}
Although generative AI has introduced new paradigms for delivering and personalizing instruction, its long-term educational impact depends on meshing these capabilities with foundational teaching frameworks. Here, we examine how the Socratic method—known for its emphasis on questioning and reflection—can be enhanced by AI-driven interactivity, and what ethical and practical considerations must be addressed to ensure that advanced technology truly serves learning.

\subsection{The Socratic Model in Modern Education}
The Socratic method, characterized by systematic questioning to stimulate critical thinking and uncover assumptions, has endured as a cornerstone of effective pedagogy \citep{PaulElder2008}. It emphasizes dialogue and inquiry over rote memorization, encouraging learners to question, reflect, and build deeper understanding.

In contemporary education, the Socratic model’s relevance has been amplified by the need for students to navigate an increasingly complex and data-rich world. Critical thinking—defined as the ability to analyze, evaluate, and synthesize information—is now recognized as a vital competency for success across disciplines \citep{Facione1990}. By empowering learners to constantly probe their own ideas, the Socratic approach dovetails well with advanced LLMs that can offer immediate, context-sensitive prompts.

However, as advocated in the concluding paragraph of Section 1, it is vital that educators maintain a guiding hand. AI systems—no matter how advanced—may pose questions or highlight misconceptions, but deep learning requires purposeful facilitation and scaffolding, which AI alone has yet to fully replicate.

\subsection{Generative AI in Facilitating Socratic Dialogue}
Generative AI, exemplified by advanced models like o3 and GPT-4, has demonstrated the capacity to emulate conversational patterns and adapt responses based on user input \citep{OpenAI2023}. These systems can replicate and even extend the Socratic method by posing probing questions, diagnosing conceptual gaps, and encouraging reflective responses. For example, AI-driven conversational tools like AutoTutor have shown success in creating interactive learning environments that mimic one-on-one tutoring \citep{Graesser2004}.

\textbf{Key Advantages} include:
\begin{itemize}
    \item \emph{Personalization at Scale}: AI systems tailor questions to the learner’s cognitive level, addressing individual needs more flexibly than traditional instruction.
    \item \emph{Immediate Feedback and Iteration}: Learners receive instant feedback, adjusting their understanding in real time.
    \item \emph{Content Diversity and Context Sensitivity}: LLMs can draw from vast repositories of knowledge, offering examples, analogies, or counterarguments that broaden learners’ perspectives \citep{Nye2014}.
\end{itemize}

\textbf{Cautions and Ethical Considerations} remain. Achieving meaningful discourse requires carefully designed question structures and follow-up prompts. Moreover, data privacy is a concern, as advanced LLMs often rely on large volumes of user data for fine-tuning. As before, it is the human educator, not merely the tool, that ultimately orchestrates the environment for genuine inquiry.

\subsection{Research and Development Framework}
\label{sec:framework}
Ensuring that generative AI in education meets genuine learning goals requires coherent frameworks and evaluation metrics. This section outlines how advanced LLMs can be integrated under evidence-based learning theories and systematically assessed, echoing the stance that powerful technology alone cannot drive learning without robust pedagogy.

\subsection{Integrative Frameworks for AI in Education}
The \emph{ALTTAI} framework (Advanced Learning Theories, Technologies, Applications, and Impacts) seeks to harmonize AI capabilities with well-established educational strategies:
\begin{itemize}
    \item \textbf{Advanced Learning Theories}: Cognitive psychology and educational neuroscience inform system design. Adaptive learning, aligning with learners’ cognitive states, significantly boosts retention \citep{ClarkMayer2016}.
    \item \textbf{State-of-the-Art Technologies}: High-performing LLMs, such as o3, can create interactive content, from domain-specific problem sets to immersive simulations \citep{Woolf2021}.
    \item \textbf{Strategic Applications of AI}: Successful deployment of advanced models depends on educators and technologists adapting to real-world contexts, including subject matter, demographic factors, and learning objectives \citep{CollinsHalverson2018}.
    \item \textbf{Proven Efficacy}: Ongoing research—like randomized controlled trials—ensures that new AI interventions deliver meaningful improvements.
\end{itemize}

By weaving these elements together, LLM-driven systems can bolster creative thinking, self-regulated learning, and mastery of content, while still requiring instructors to anchor them in best-practice pedagogy.

\subsection{Metrics for Evaluating Impact}
A well-rounded evaluation of AI integration in education encompasses:
\begin{enumerate}
    \item \textbf{Learning Outcomes}: Measures such as standardized test scores, mastery of competencies, and performance in open-ended tasks.
    \item \textbf{Engagement and Retention}: Time-on-task, dropout rates, and learner motivation, often assessed through interactive logs and surveys.
    \item \textbf{Equity and Accessibility}: Data on device availability and whether AI tools accommodate linguistic and cultural diversity.
    \item \textbf{Ethical Compliance}: Safeguards around student privacy and fairness in decision-making, ensuring no demographic biases are perpetuated.
    \item \textbf{Learner Autonomy}: Evidence of deep inquiry, critical thinking, and self-reflection rather than superficial or formulaic responses.
\end{enumerate}

As emphasized earlier, robust instructional design and teacher-mediated scaffolding greatly influence whether advanced AI systems truly elevate learning or merely automate certain tasks.

\section{AutoTutor: Legacy, Challenges, and Unfulfilled Aspirations}
\label{sec:autotutor}
AutoTutor was among the first Intelligent Tutoring Systems to use adaptive conversational agents, demonstrating the potential of interactive dialogue in advancing learning. This section revisits its core achievements, focusing on how AutoTutor pioneered expectation-misconception tailored (EMT) pedagogy. We then address the constraints that limited its scalability and personalization, and reflect on the more advanced horizons enabled by Large Language Models (LLMs).

\subsection{EMT System within AutoTutor}
\label{subsec:emt-autotutor}
The EMT (Expectation-Misconception Tailored) system within AutoTutor is designed to engage with students interactively by tailoring prompts, hints, and feedback based on their answers. Figure~\ref{fig:emt-pedagogy} illustrates the workflow:

\begin{figure}[htbp]
    \centering
    \includegraphics[width=0.85\textwidth]{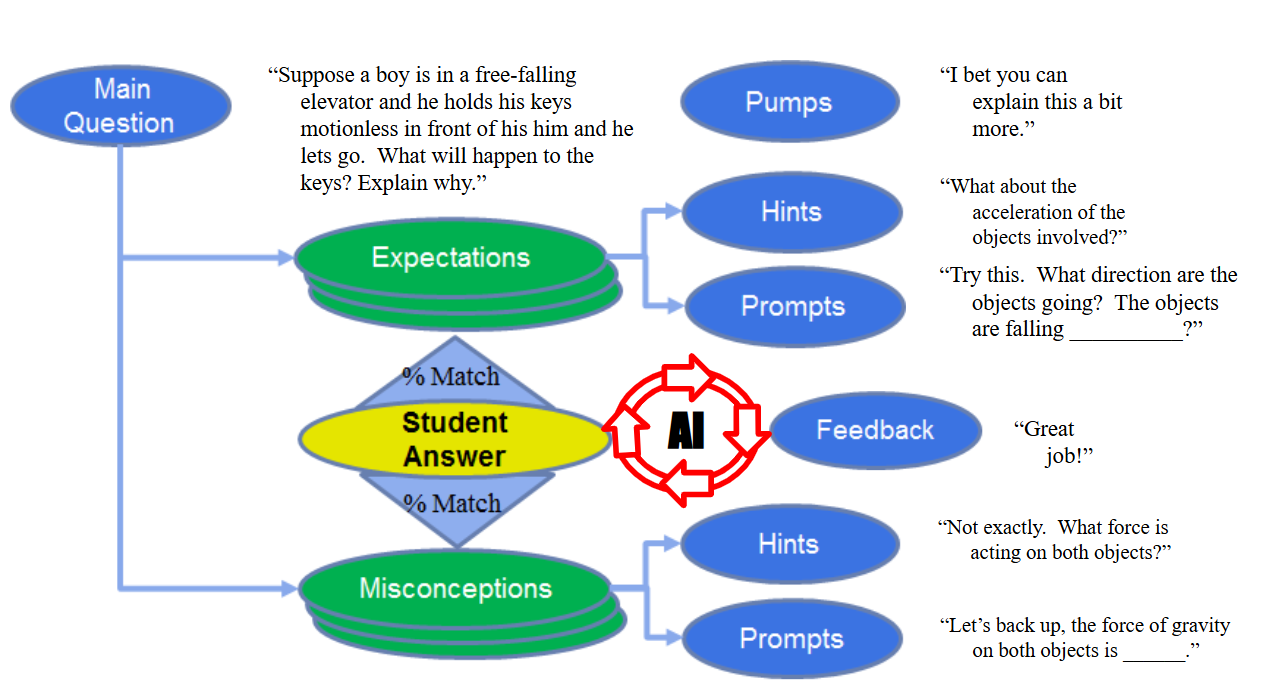}
    \caption{The EMT framework in AutoTutor, illustrating how the system compares a student’s answer to expectations and misconceptions, then tailors hints, prompts, and feedback accordingly.}
    \label{fig:emt-pedagogy}
\end{figure}

\noindent\textbf{Main Question Presentation.}
A primary question is presented to the student (e.g., “Suppose a boy is in a free-falling elevator...”). This question sets the context and invites the student’s reasoning.

\noindent\textbf{Student Answer Evaluation.}
The student’s response is evaluated against two frameworks:
\begin{itemize}
    \item \textbf{Expectations:} Correct or partially correct conceptual understandings of the topic.
    \item \textbf{Misconceptions:} Common incorrect understandings or errors related to the concept.
\end{itemize}
The system measures the percent match of the student’s response to both expectations and misconceptions.

\noindent\textbf{Dynamic AI-Driven Feedback Loop.}
Based on the match percentage, the system determines the next step:
\begin{itemize}
    \item If the response aligns well with expectations, it provides feedback that encourages further elaboration (e.g., “Great job!” or “I bet you can explain this a bit more.”).
    \item If the response indicates misconceptions, the system delivers tailored hints or prompts to guide the student toward the correct understanding.
\end{itemize}
Examples include:
\begin{itemize}
    \item \emph{Hints:} “What about the acceleration of the objects involved?”
    \item \emph{Prompts:} “Let’s back up, the force of gravity on both objects is \_\_\_\_\_.”
\end{itemize}

\noindent\textbf{Iterative Learning Process.}
The AI system engages in an iterative loop with the student, continually evaluating updated answers, matching them to the frameworks, and adjusting its feedback dynamically. The focus is to help the student gradually refine their understanding by addressing misconceptions and reinforcing correct knowledge.

\noindent\textbf{Scaffolding Techniques.}
AutoTutor strategically uses hints, prompts, and feedback to scaffold learning:
\begin{itemize}
    \item \textbf{Hints:} Subtle cues that nudge the student’s thinking in a productive direction.
    \item \textbf{Prompts:} More directive statements or questions that break down reasoning steps for the student.
    \item \textbf{Feedback:} Affirmations or corrective suggestions that encourage deeper engagement.
\end{itemize}
Overall, EMT is a structured system designed to personalize the learning experience by aligning feedback with each student’s specific needs, ultimately improving conceptual understanding.

\subsection{Challenges in AutoTutor’s Implementation}
Despite AutoTutor’s innovation, several technological and practical challenges constrained its scalability, personalization, and broader educational impact:

\begin{itemize}
    \item \textbf{NLP Limitations:} Reliance on earlier techniques such as Latent Semantic Analysis (LSA) led to difficulties in parsing more creative or complex answers.
    \item \textbf{Static Content Framework:} The system depended heavily on scripted misconceptions and feedback, limiting adaptability to new topics.
    \item \textbf{Scalability:} Expanding AutoTutor to many subject areas required significant effort in domain modeling and content authoring.
\end{itemize}

\noindent
Building upon these early foundations, subsequent generations of tutoring systems seek to leverage more powerful LLMs and advanced adaptive strategies to reach the level of personalized interaction initially envisioned by AutoTutor’s EMT framework.

\subsection{Unfulfilled Pedagogical Aspirations}
\label{subsec:unfulfilled-auto}
\begin{figure}[htbp]
    \centering
    \includegraphics[width=0.9\textwidth]{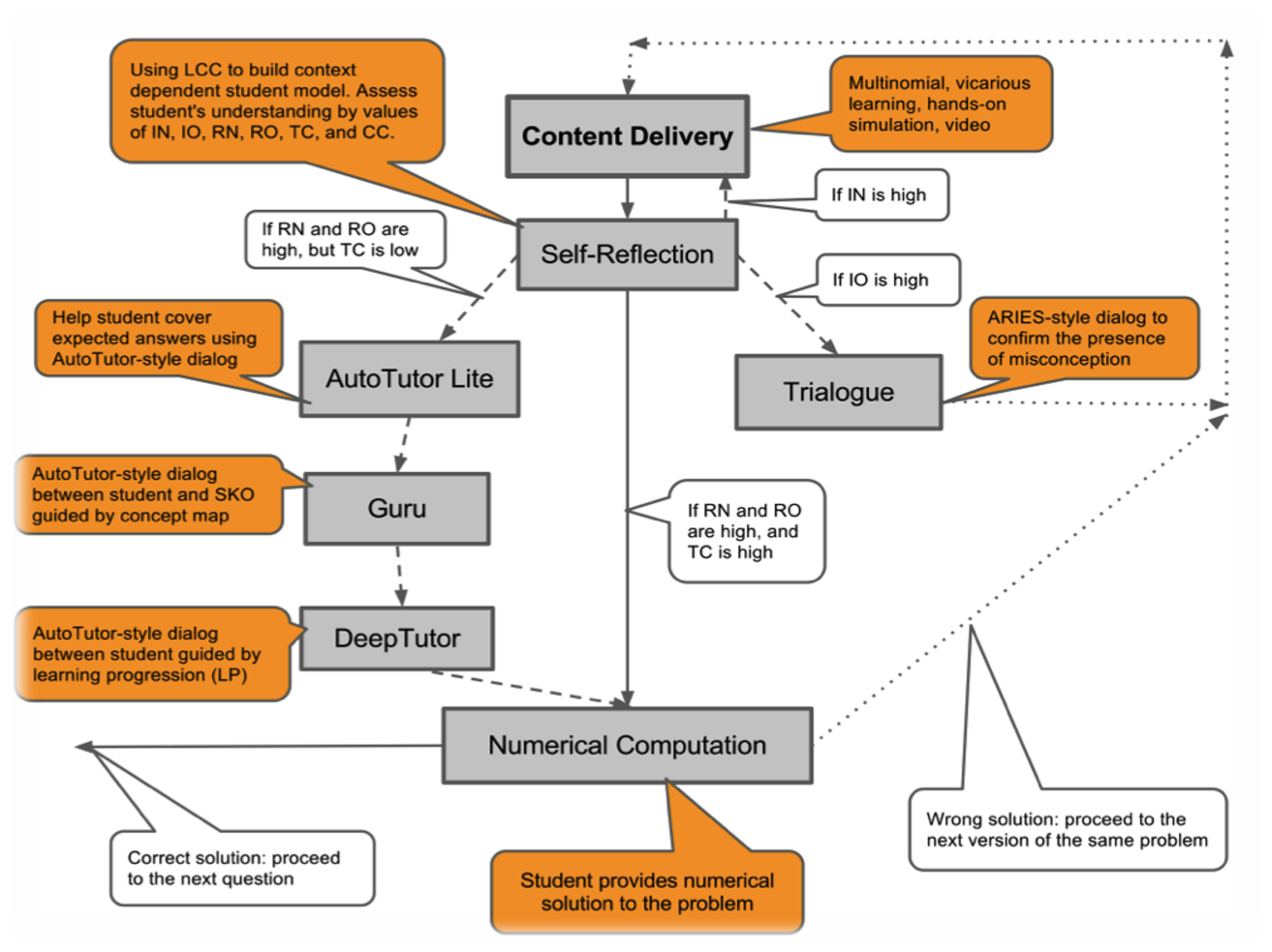} 
    \caption{High-level design of the ONR STEM Grand Challenge application, integrating multiple tutoring modalities and the Learner's Characteristics Curve (LCC) to create an adaptive learning environment. }
    \label{fig:onrSTEMDesign}
\end{figure}

In addition to the early ambitions outlined previously, AutoTutor’s involvement in the Office of Naval Research (ONR) STEM Grand Challenge further illustrates the scope of aspirations and challenges encountered during large-scale implementation. Figure~\ref{fig:onrSTEMDesign} depicts the high-level design of the application, showcasing how multiple tutoring modalities and a Learner’s Characteristics Curve (LCC) were intended to create a fully adaptive learning environment.

AutoTutor served as the cornerstone technology behind the ONR STEM Grand Challenge, an ambitious initiative to revolutionize STEM education through conversation-based intelligent tutoring. This challenge aimed to implement AutoTutor across a variety of STEM disciplines, leveraging adaptive features to meet diverse student needs. The diagram in Figure~\ref{fig:onrSTEMDesign} shows the interplay of multiple tutoring modalities alongside the Learner’s Characteristics Curve (LCC), all designed to create an immersive and personalized educational experience.

At its core, AutoTutor employs natural language dialogue to guide students through problem-solving and conceptual discussions, simulating human tutor interactions to promote deep learning. Its early successes in STEM domains provided the impetus for the ONR STEM Grand Challenge, which hoped to use AutoTutor’s conversation-based methods to improve outcomes for students in science, technology, engineering, and mathematics. The vision was to embed AutoTutor’s conversational strengths into a robust, multi-modal tutoring platform that could adapt to a wide range of student proficiencies and topics.

A key mechanism for AutoTutor’s adaptivity in the ONR STEM Grand Challenge was the Learner’s Characteristics Curve (LCC), originally developed by Hu, Morrison, and Cai \citep{Hu2013GIFT1}. The LCC decomposes student responses into four key components:
\begin{itemize}
    \item \textbf{Relevant-New (RN):} Information pertinent to the expected answer that the student has not yet introduced.
    \item \textbf{Irrelevant-New (IN):} New information the student provides that does not align with the expected answer.
    \item \textbf{Relevant-Old (RO):} Previously discussed but relevant information reiterated by the student.
    \item \textbf{Irrelevant-Old (IO):} Previously mentioned, but irrelevant information repeated in the current context.
\end{itemize}
By parsing learner inputs into these categories, the system can dynamically tailor feedback, focusing on unmet learning objectives while gently correcting misalignments. In principle, this level of fine-grained adaptivity was central to advancing AutoTutor beyond static conversation scripts, ensuring that each learner’s unique path through STEM problem-solving was recognized and supported.

To accommodate a wide range of learners and contexts, the ONR STEM Grand Challenge framework included multiple tutoring modalities:
\begin{enumerate}
    \item \textbf{AutoTutor Lite:} Provides hint-only interactions. The minimal scaffolding ensures learners remain active in proposing solutions but still benefit from carefully timed guidance.
    \item \textbf{GuruTutor:} Relies on concept maps to facilitate more structured dialogues, helping students form coherent mental models of relationships among ideas.
    \item \textbf{DeepTutor:} Focuses on deeper, adaptive dialogues targeted toward individual learning progressions. It aims to personalize instruction at a finer granularity than other modalities.
    \item \textbf{ARIES (Trialogue Interaction):} Engages a student, a tutor, and a virtual peer, simulating multi-party dialogues to identify and correct misconceptions more dynamically.
\end{enumerate}
This multi-modal approach was designed so that learners could shift between modalities as their needs evolved, reflecting the LCC’s assessment of their contributions.

Despite its innovative design, the real-world application of AutoTutor in this expansive STEM context encountered notable obstacles:
\begin{itemize}
    \item \textbf{Integration Complexity:} Coordinating multiple tutoring modalities and managing transitions among them (e.g., from AutoTutor Lite to DeepTutor) demanded extensive design and engineering effort.
    \item \textbf{Scalability Issues:} Supporting numerous STEM subdomains, each with unique terminology and problem structures, required significant resources in terms of authoring scripts, generating domain-relevant feedback, and maintaining robust NLP performance.
    \item \textbf{Technological Limitations:} While AutoTutor excelled in certain forms of natural language dialogue, ensuring it accurately understood and responded to STEM-specific questions across a wide range of topics posed ongoing technical challenges.
\end{itemize}
These implementation hurdles underscored the gap between the aspirational goals of the ONR STEM Grand Challenge and the practical complexities of deploying an advanced intelligent tutoring system at scale.

The ONR STEM Grand Challenge’s deployment of AutoTutor highlights both the system’s promise and its unfulfilled aspirations. Although the initiative aimed to harness conversation-based tutoring to transform STEM education, complexities related to integration, scalability, and technical performance limited the extent to which these ambitions were fully realized. Nonetheless, valuable lessons emerged about the intricacies of developing large-scale, adaptive solutions for STEM learning. Future designs can build upon these insights, refining not only AutoTutor’s core capabilities but also the interplay of multiple tutoring modalities and the LCC, ultimately moving the field closer to realizing the vision of truly adaptive, high-impact intelligent tutoring systems.

Although these unfulfilled goals illustrate the hurdles faced by early ITS, they also hint at the innovations that would later redefine adaptive tutoring. The next section introduces the \emph{Socratic Playground for Learning}, a next-generation conversation-based ITS built on modern transformer-based models, designed to address many of AutoTutor’s constraints. We also offer a detailed prompt approach that shows how carefully structured JSON instructions can guide learners toward deeper reflection and self-assessment, illustrating how AI tutors can evolve beyond static templates—all while keeping pedagogy at the forefront.

\section{The Socratic Playground for Learning}
\label{sec:playground}
Building on AutoTutor’s achievements yet mindful of its limitations, the \emph{Socratic Playground for Learning} is a conversation-based Intelligent Tutoring System (ITS) that leverages state-of-the-art transformer-based NLP and robust pedagogical strategies. Unlike earlier ITS architectures, it interprets nuanced learner inputs with high accuracy and responds dynamically, generating new scenarios and feedback in real time. This system aims to scale across diverse domains with minimal manual scripting, thanks to continuous profiling of student performance and adaptive scaffolding mechanisms.

A pilot implementation of the Socratic Playground for Learning (SPL), powered by GPT-4, demonstrated significant improvements in tutoring interactions and dialogue-based ITS functionalities. \citep{zhang2024splsocraticplaygroundlearning,electronics13244876}  The system employs the Socratic teaching method to foster critical thinking among learners, generating specific learning scenarios and facilitating efficient multi-turn tutoring dialogues through extensive prompt engineering. These findings suggest that integrating large language models like GPT-4 with the Socratic teaching method can significantly enhance the effectiveness of dialogue-based ITSs in personalized learning.

\subsection{Core Design Innovations}
\begin{itemize}
    \item \textbf{Advanced NLP Precision}: Transformer models interpret complex, partially correct, or ambiguous learner inputs far more flexibly than older statistical methods like LSA.
    \item \textbf{Dynamic Content Generation}: The Playground can generate new scenarios, questions, and feedback on the fly, providing robust adaptability.
    \item \textbf{Scalability Across Domains}: By training on domain-specific corpora, it expands into new subject areas without extensive manual re-scripting.
    \item \textbf{Adaptive Personalization}: Continuous profiling of student performance, engagement, and emotional cues enables fine-grained scaffolding and motivational support.
\end{itemize}

\subsection{Five Interactive Modes and Their Logic Progression}

The five interactive modes in the Socratic Playground represent innovative strategies for adaptive learning, tailored to diverse educational needs and contexts. These modes collectively aim to personalize instruction, encourage critical thinking, and foster deeper conceptual understanding through advanced AI capabilities.
\begin{itemize}
    \item \textbf{Assessment Mode}: Learners begin with adaptive quizzes or tasks designed to identify their understanding, misconceptions, and confidence levels. The system uses this information to recommend the appropriate next steps in their learning journey.
    \item \textbf{Tutoring Mode}: This mode offers individualized, conversational instruction. It pinpoints misconceptions, provides tailored hints, and delivers feedback to guide learners toward refining their knowledge.
    \item \textbf{Vicarious Mode}: Learners observe virtual discussions among multiple participants, gaining insights into problem-solving strategies and discourse skills. This mode promotes learning through observation while minimizing performance pressure.
    \item \textbf{Gaming Mode}: Through gamified challenges, learners engage with concepts in an interactive, competitive, or cooperative format. This mode enhances engagement and reinforces knowledge through active participation.
    \item \textbf{Teachable Agent Mode}: Learners teach a simulated struggling student, explaining concepts, addressing misconceptions, and providing feedback. This role-reversal method strengthens mastery and metacognitive skills.
\end{itemize}

These modes leverage AI-driven scaffolding and feedback to cater to varying learner needs, seamlessly blending technology and pedagogy for impactful education.

The five interactive modes in the Socratic Playground are structured to guide learners through a logical progression tailored to their needs. Learning begins in \textbf{Assessment Mode}, where adaptive quizzes evaluate understanding, confidence, and misconceptions. Based on these results, the system directs learners to the appropriate next step. For foundational gaps, \textbf{Tutoring Mode} provides targeted, one-on-one guidance to clarify misconceptions and refine knowledge. When learners are ready for independent application, they engage with \textbf{Gaming Mode}, using competitive or cooperative challenges to reinforce and apply concepts. Alternatively, \textbf{Vicarious Mode} offers observation-based learning, where learners watch structured discussions, reducing pressure while absorbing strategies. At the highest level, \textbf{Teachable Agent Mode} tasks learners with teaching a simulated student, enhancing mastery through explanation and feedback. This iterative framework allows learners to cycle back to reassessment as needed, ensuring a dynamic, adaptive pathway that fosters deeper understanding and sustained engagement.

\subsection{Anatomy of an AutoTutor-LLM Prompt}
In the Appendix, we present a JSON-based prompt approach that showcases how carefully structured instructions let the system track “Expectations,” “Misconceptions,” and partial overlaps. This architecture helps the AI calibrate feedback to the user’s evolving input, ensuring that each tutoring turn is purposeful and fosters deeper conceptual clarity.

\section{Future Directions: A Narrative of Emerging Possibilities}
\label{sec:future}
While the Socratic Playground demonstrates how advanced NLP and thoughtful pedagogy can converge, the potential of Intelligent Tutoring Systems (ITS) extends well beyond individualized content delivery. Drawing on guidelines from the \emph{Design Recommendations for Intelligent Tutoring Systems} and the Generalized Intelligent Framework for Tutoring (GIFT), researchers are now looking toward learning ecosystems that integrate cognitive, emotional, and social dimensions—always mindful that the most sophisticated technologies succeed only when paired with robust instructional strategies.

\subsection{Team Tutoring and Self-Improving Systems}
A promising trajectory involves \emph{team tutoring} \citep{Hu2018GIF1}, where next-generation ITS orchestrate collaborative problem-solving among groups of learners. AI can moderate discussions, distribute tasks equitably, and analyze group dynamics to ensure balanced participation. In parallel, \emph{self-improving} adaptive systems \citep{Hu2019GIF1refs} continuously refine their pedagogical logic based on large-scale learner data. Self-improving systems\citep{Tong2019GIF1refs}, employ iterative learning loops to refine instructional strategies over time. By simulating diverse virtual learners and analyzing large-scale interaction data, these systems enable dynamic adjustments to pedagogical methods, ensuring alignment with evolving learner needs and group dynamics. This capability bridges individual learning and collaborative environments, driving continuous improvement in intelligent tutoring systems. By simulating diverse “virtual learners,” these systems can predict the efficacy of new instructional interventions before rolling them out in real classrooms.

\subsection{Metacognitive Dashboards and Behavioral Analytics}
Another trend is the development of \emph{metacognitive dashboards}, which empower learners to visualize their progress, set personalized goals, and monitor study habits. Combined with behavioral analytics, these dashboards can identify engagement dips or cognitive overload, prompting timely interventions. The overarching objective is not merely to “push” more content but to shape an environment where learners self-regulate, reflect, and engage deeply.

\subsection{Equitable Access and Global Collaboration}
Equitable access remains a core concern, especially as advanced AI tools often assume robust digital infrastructures. Policymakers, technologists, and educators must collaborate to ensure connectivity, affordability, and inclusive design. Without such initiatives, powerful LLMs risk widening existing educational divides rather than narrowing them.

\subsection*{Echoing Our Foundational Insights}
\noindent
Ultimately, the trajectory of AI in education will not be dictated by technology alone. The parallels between LLMs and human cognition provide a compelling reason to harness conversation-based tutoring, especially since state-of-the-art models increasingly surpass human capabilities in specific tasks. Yet, the indispensable factor remains pedagogy: advanced models can yield extraordinary results only when thoughtfully integrated into frameworks that emphasize learner engagement, reflection, and growth. As with educators’ use of new tools, more technology does not guarantee better instruction—what truly matters is a measured, strategic approach. This perspective, initially introduced in Section~\ref{sec:intro}, is reaffirmed here to illustrate that transformative learning arises when powerful AI meets well-designed teaching approaches.

\appendix

\section{Five Pedagogical Modes of Socratic Playground for Learning}

In the Socratic Playground, a learner’s journey typically begins with \emph{Assessment Mode}, then advances (or returns) to more demanding modes as needed:
\subsection{Assessment Mode (Self-Assessment)}
\label{sec:assessment-mode}
This mode is the entry point for any new concept or skill. Learners attempt an adaptive quiz or problem set tailored to gauge both their understanding and metacognitive factors such as confidence. Depending on the results, the Socratic Playground recommends the next step—either more practice, Tutoring Mode, or advanced modes if mastery is apparent.

\begin{figure}[htbp]
    \centering
    \includegraphics[width=0.65\textwidth]{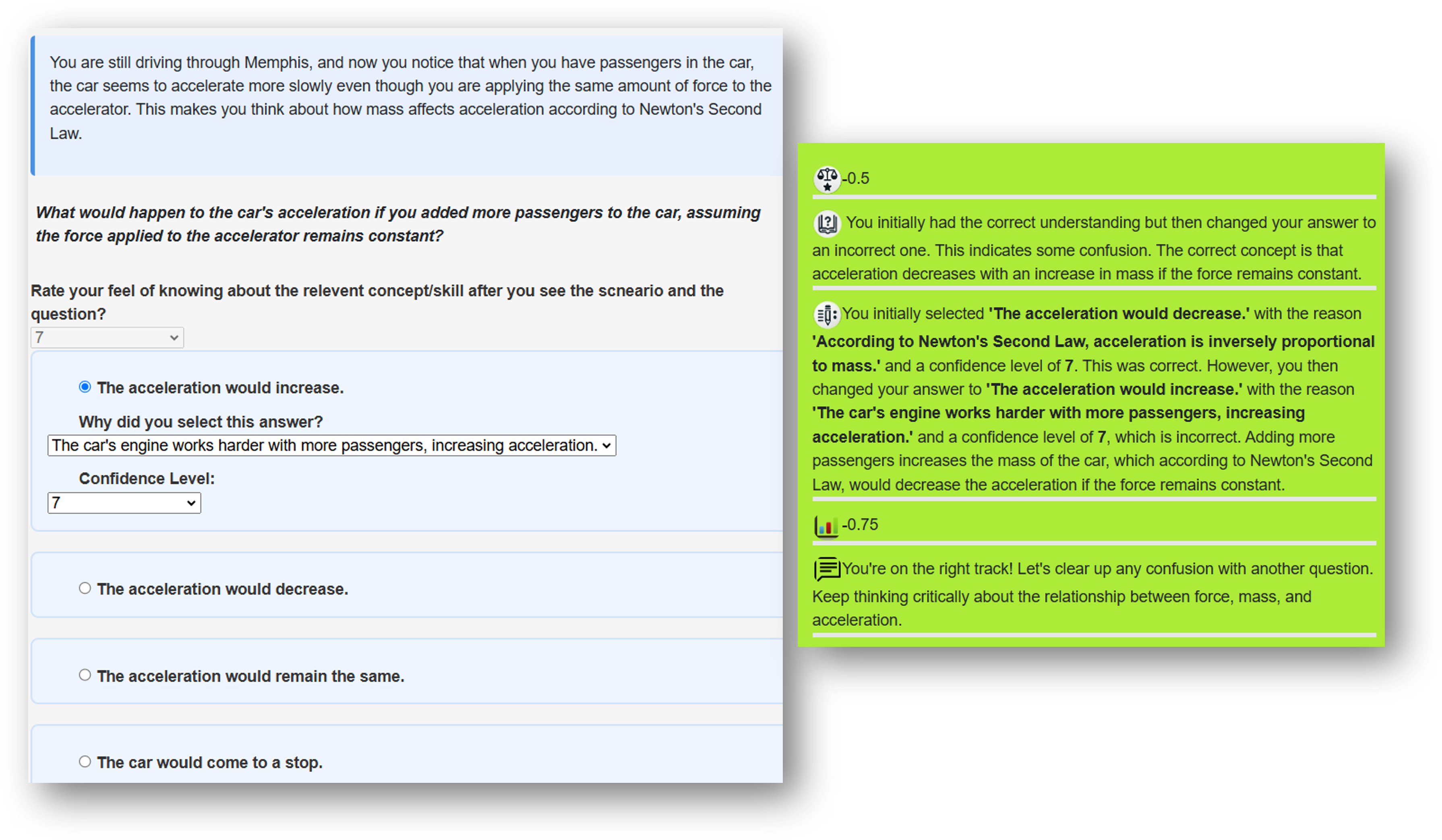}
    \caption{An adaptive self-assessment interaction showing both the learner’s question/answer panel (left) and context-sensitive feedback (right).}
    \label{fig:newton-selfassess}
\end{figure}

\noindent\textbf{Example Interaction.}
Figure~\ref{fig:newton-selfassess} captures an interaction with an adaptive self-assessment tool designed to evaluate and guide the learner’s understanding of Newton’s Second Law of Motion. The learner is given a real-world scenario about a car’s acceleration when additional passengers (i.e., more mass) are added, while force remains constant.

\medskip
\noindent\textbf{Context (Left Panel).}
\begin{enumerate}
    \item \emph{Scenario and Question:} 
    The scenario frames how acceleration depends on force and mass, asking how acceleration changes if more passengers (mass) are added while the force remains unchanged.
    \item \emph{Learner’s Interaction:} 
    Initially, the learner correctly selects “The acceleration would decrease” with the reasoning: 
    \emph{“According to Newton’s Second Law, acceleration is inversely proportional to mass.”}
    Later, they switch to an incorrect choice: “The acceleration would increase,” mistakenly assuming the engine compensates by working harder, despite the force being specified as constant.
    \item \emph{Confidence Level:} 
    The learner expresses high confidence (7/7) for both the initial correct and subsequent incorrect answer, suggesting either confusion or overconfidence.
\end{enumerate}

\noindent\textbf{Feedback (Right Panel).}
\begin{itemize}
    \item \emph{Recognition of Error:} 
    The system notes that the learner’s first answer was conceptually correct before switching to an incorrect response, which contradicts Newton’s Second Law.
    \item \emph{Conceptual Guidance:} 
    It explains that adding mass reduces acceleration when force remains constant.
    \item \emph{Encouragement and Prompting:} 
    The feedback encourages the learner to revisit the relationship between force, mass, and acceleration, reinforcing the correct principle.
\end{itemize}

\noindent\textbf{Analysis of the Interaction.}
\begin{enumerate}
    \item \emph{Cognitive Confusion:} 
    The learner conflates the idea of an engine “working harder” with the concept of a truly constant force.
    \item \emph{High Confidence in Error:} 
    This highlights that confidence alone is not a reliable indicator of correctness, underscoring the importance of detailed feedback.
    \item \emph{Adaptive Design of Feedback:} 
    The system tailors hints and prompts directly to the learner’s misconceptions, clarifying the misunderstood physics principle.
\end{enumerate}

\noindent\textbf{Takeaways.}
This example illustrates how adaptive self-assessment tools can diagnose and address misunderstandings in real time. By contextualizing physics concepts, capturing confidence levels, and offering targeted feedback, learners are nudged toward deeper, more accurate reasoning—even when they start out with incorrect (yet confidently held) notions. Through iterative feedback loops, the system aligns learner intuition with scientific principles, exemplifying the importance of well-designed, self-assessment-oriented pedagogy.

\subsection{Tutoring Mode (One-on-One Guidance)}
\label{sec:tutoring-mode}
\begin{figure}[htbp]
    \centering
    \includegraphics[width=0.7\textwidth]{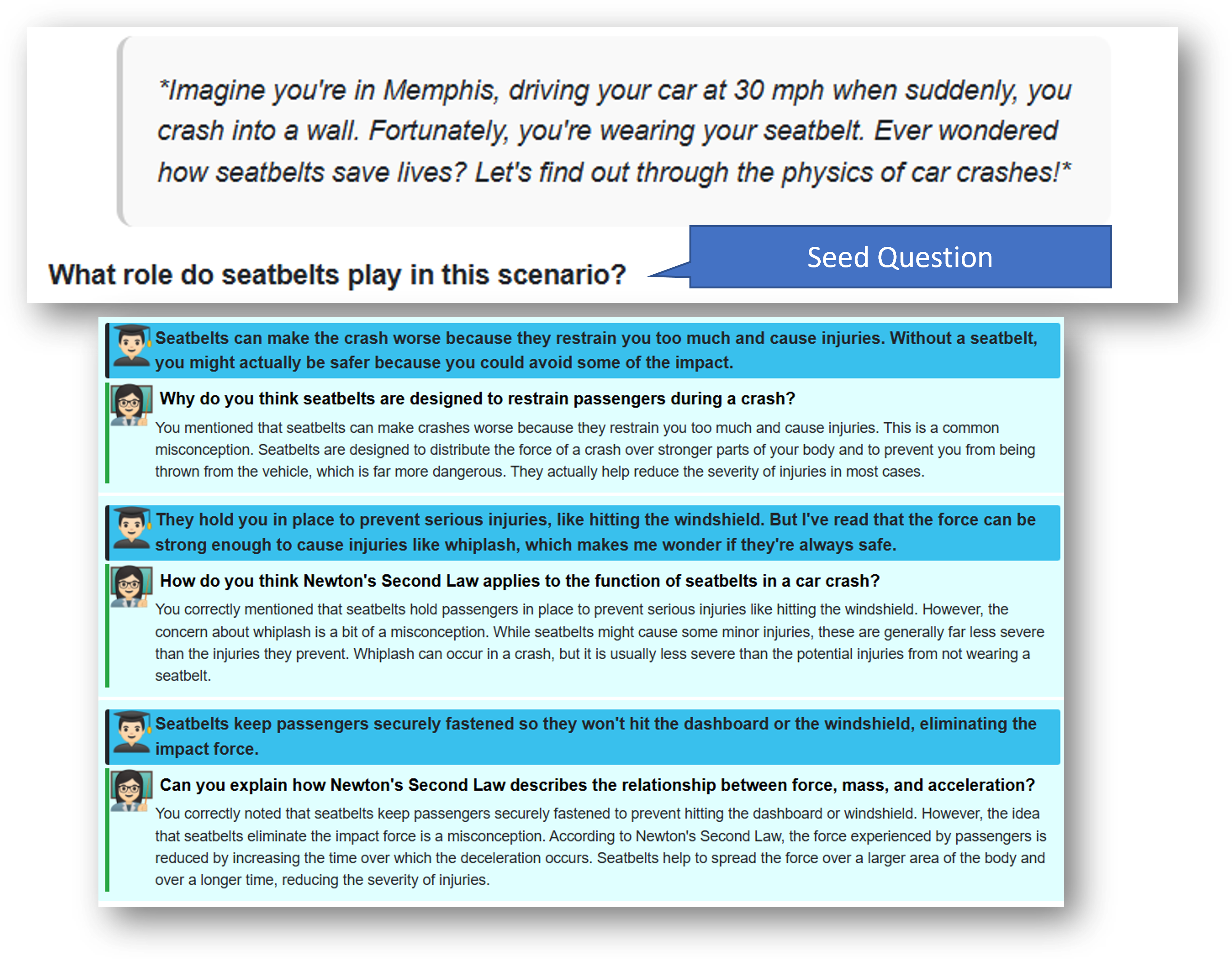}
    \caption{A seatbelt-themed tutoring scenario. The learner initially expresses misconceptions about seatbelts “causing more harm,” prompting the tutor to provide clarifications related to force distribution, Newton’s Second Law, and injury prevention.}
    \label{fig:tutoring-seatbelts}
\end{figure}

Tutoring Mode focuses on individualized, interactive instruction. In this setup, an AI tutor or human–AI collaboration pinpoints misconceptions, delivers targeted hints or prompts, and supports learners in refining their conceptual understanding. Figure~\ref{fig:tutoring-seatbelts} illustrates a conversation scenario in which seatbelts and car crashes serve as the learning context.

\noindent
\textbf{Context and Approach.} In the scenario, the learner is asked, “What role do seatbelts play?” as they imagine crashing into a wall at 30 mph. The tutoring system then engages the learner with follow-up questions, uncovering misconceptions—such as believing that seatbelts might \emph{cause} injuries—and guiding them through structured feedback that references Newton’s Second Law (i.e., force = mass \(\times\) acceleration). A key emphasis is on how seatbelts spread the impact force over a larger area and lengthen the deceleration time to reduce injury.

\medskip

\noindent
\textbf{Illustrated Steps in the Dialogue.}
\begin{itemize}
    \item \textbf{Learner’s Responses (blue boxes):} These reflect the learner’s evolving thoughts. Early answers hint at misconceptions (“Seatbelts can make the crash worse…”). 
    \item \textbf{Tutor’s Probing Questions:} The tutor probes deeper, asking, “Why do you think seatbelts are designed to restrain passengers?” to prompt critical reflection on the function of seatbelts.
    \item \textbf{Tutor’s Feedback and Corrections:} In green text (or another highlight), the system clarifies how seatbelts distribute force more safely, notes common misunderstandings (e.g., “Seatbelts might cause whiplash”), and connects the conversation back to Newton’s Second Law. 
\end{itemize}

\noindent
\textbf{Pedagogical Strategies at Work.}
\begin{enumerate}
    \item \emph{Contextualization:} By situating learning in a realistic scenario—car crashes at typical speeds—students more readily appreciate the relevance of physics principles.
    \item \emph{Misconception Correction:} The tutor systematically addresses ideas like “seatbelts cause more harm than good” or “seatbelts completely eliminate force” by highlighting factual corrections.
    \item \emph{Socratic Inquiry:} Questions such as “How do you think Newton’s Second Law applies?” guide learners toward self-discovery of correct concepts, rather than simply telling them the right answer.
    \item \emph{Scaffolding and Feedback:} The conversation begins with broad seatbelt queries, then narrows to specific physics principles (deceleration, force distribution). The tutor’s comments build on each learner statement, reinforcing correct ideas and gently correcting errors.
\end{enumerate}

\noindent
\textbf{Outcome.} This one-on-one guidance mode not only highlights key physics concepts (e.g., how seatbelts mitigate injury by changing the time–force relationship) but also personalizes instruction by responding immediately to the learner’s misconceptions. Over multiple tutor–learner exchanges, the system helps students reconcile intuitive misunderstandings with scientifically sound principles, illustrating how seatbelt usage aligns with core physics laws and substantially increases crash safety.

\noindent
\textbf{Learner’s Characteristics Curve (LCC) Analysis.}
Another critical part of this tutoring approach involves tracking the learner’s contributions according to the LCC framework \citep{Hu2013GIFT1}—categorizing each statement by its alignment with previously introduced expectations or misconceptions. Figure~\ref{fig:lcc-analysis} illustrates how the system scores and records each learner turn:

\begin{figure}[htbp]
    \centering
    \includegraphics[width=0.85\textwidth]{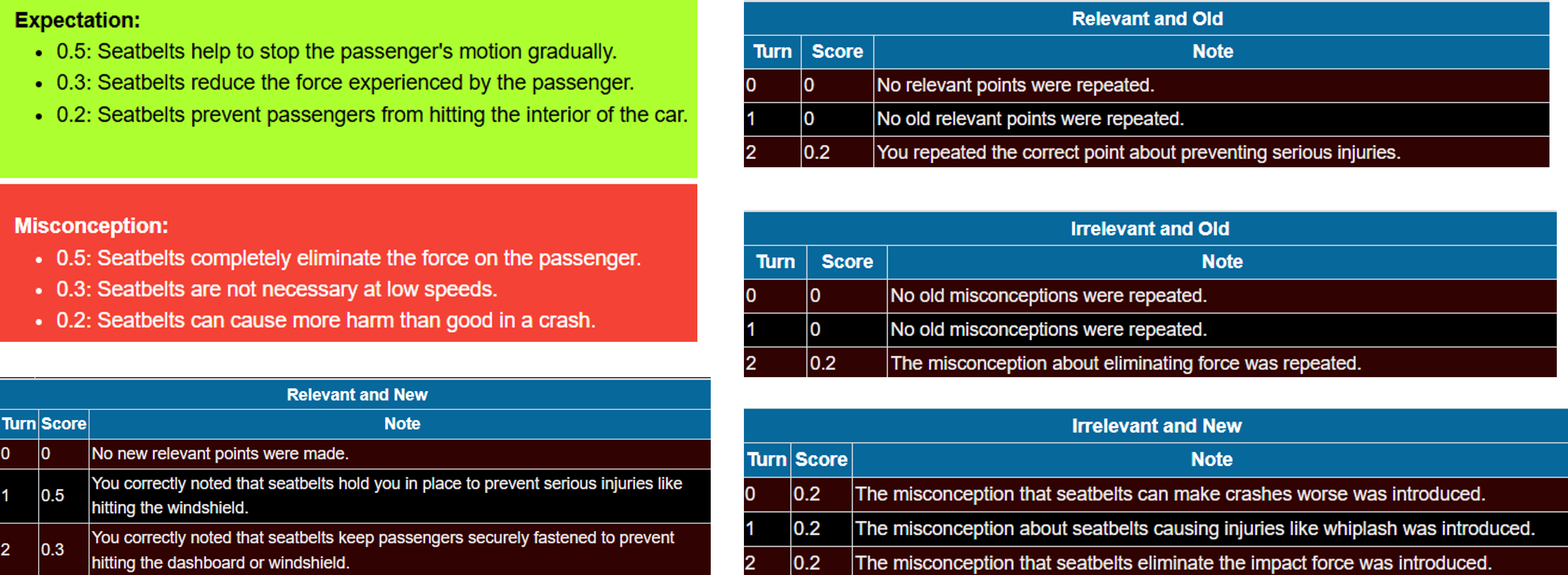}
    \caption{LCC analysis of each learner contribution, showing scores for Relevant/Irrelevant and New/Old points, as well as any repeated misconceptions. }
    \label{fig:lcc-analysis}
\end{figure}

\noindent
\textbf{How the LCC Analysis Works.}
\begin{itemize}
    \item \emph{Expectations vs. Misconceptions:}
    At the top of Figure~\ref{fig:lcc-analysis}, the system lays out which ideas are correct (expectations) and which are incorrect (misconceptions). Each has a weight (e.g., 0.5, 0.3, 0.2) reflecting its relative importance or level of detail.
    \item \emph{Turn-by-Turn Scoring:}
    For each learner turn (rows labeled 0, 1, 2, etc.), the system checks whether new or repeated ideas match expectations or misconceptions.
    \begin{itemize}
        \item \emph{Relevant and Old/New:} Tracks how many correct points are repeated or newly introduced.
        \item \emph{Irrelevant and Old/New:} Identifies misconceptions that resurface or arise for the first time.
    \end{itemize}
    \item \emph{Dynamic Feedback:}
    Based on these categories, the tutor can redirect the conversation or reinforce correct points. For instance, it might award a partial score if a learner repeats a partially correct statement or carefully correct a repeated misconception about “seatbelts causing more harm” if it appears multiple times.
\end{itemize}

\noindent
\textbf{Integrating LCC with Tutoring.}
This LCC-based analysis provides transparency about how each learner statement is interpreted, making it easier for educators or system designers to see:
\begin{itemize}
    \item \emph{Which misconceptions persist over multiple turns.}
    \item \emph{How well learners consolidate correct ideas.}
    \item \emph{Where additional scaffolding or deeper inquiry might be necessary.}
\end{itemize}
Together, the seatbelt scenario and LCC evaluation illustrate how a robust tutoring system not only engages learners in dialogue but also systematically tracks and responds to each contribution, ensuring misconceptions are addressed promptly and correct understandings are strengthened.

Overall, this one-on-one guidance mode reflects how an AI tutor can seamlessly blend domain knowledge, real-world examples, and reflective questioning to solidify physics concepts. Particularly in the realm of seatbelt physics, the conversation underscores why seatbelts are essential—showcasing how small changes in deceleration time significantly reduce forces acting on passengers and prevent severe injuries.
\subsection{Vicarious Mode (Observational Learning)}
\label{sec:vicarious-mode}
\begin{figure}[htbp]
    \centering
    \includegraphics[width=0.8\textwidth]{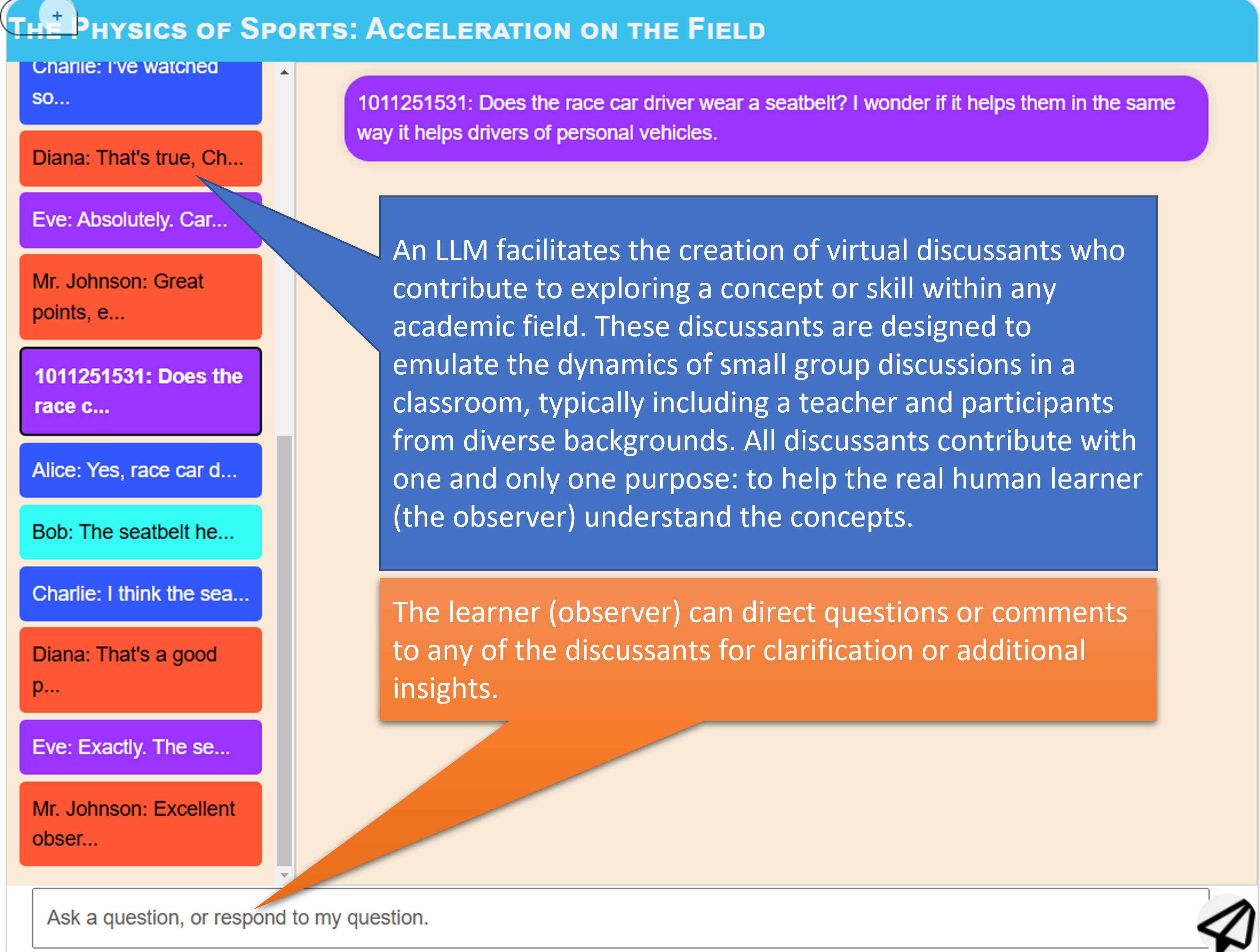}
    \caption{A Vicarious Mode interface with virtual discussants (Alice, Bob, Charlie, Diana, Eve, Mr. Johnson) conversing about sports physics. The learner (observer) occasionally poses questions, while the LLM orchestrates responses. }
    \label{fig:vicarious-interface-new}
\end{figure}

In Vicarious Mode, the learner adopts the role of a silent observer, watching a structured group conversation unfold without being required to actively participate. Figure~\ref{fig:vicarious-interface-new} presents an interface where multiple virtual discussants (Alice, Bob, Charlie, Diana, Eve, Mr. Johnson) explore the topic of “acceleration on the field”—a physics-focused look at seatbelts and racing scenarios, among other ideas.

\noindent
\textbf{Context and Design.}
\begin{itemize}
    \item \emph{Multiple Virtual Discussants:} Each participant, from Alice to Mr. Johnson, brings a unique perspective—some more physics-savvy, others with hands-on sport experience.
    \item \emph{Observer Role:} The learner watches the dialogue progress, interjecting with occasional questions or prompts (e.g., “Does the race car driver wear a seatbelt? I wonder if it helps them in the same way it helps drivers of personal vehicles.”).
    \item \emph{LLM-Facilitated Interaction:} An underlying LLM ensures each discussant stays on topic, clarifies misunderstandings, and continually refocuses on the learning objective: applying physics concepts (like force or acceleration) to real-world contexts in racing and other sports.
\end{itemize}

\noindent
\textbf{Illustrative Flow of Conversation.}
\begin{enumerate}
    \item \textbf{Scenario Introduction:} The virtual environment introduces a scenario related to seatbelts, velocity, or Newton’s laws.
    \item \textbf{Group Discussion:} Alice might remark on how seatbelts distribute force, Bob might mention a misconception (e.g., “The seatbelt could make it worse”), and Charlie or Diana might correct that misconception by referencing acceleration formulas.
    \item \textbf{Observer’s Questions:} The learner can post clarifying questions—like, “How does mass factor into the driver’s acceleration in a crash?”—and see how the group responds.
\end{enumerate}

\noindent
\textbf{Pedagogical Advantages of Vicarious Mode.}
\begin{itemize}
    \item \emph{Reducing Performance Pressure:} Since the observer does not have to speak every turn, they can learn by quietly observing how experts or peers negotiate meaning, resolve contradictions, and refine ideas.
    \item \emph{Demonstrating Discourse Skills:} Each discussant models respectful academic discourse, offering or disputing claims in a structured manner that helps the observer see what quality exchanges look like.
    \item \emph{Addressing Misconceptions Indirectly:} Bob’s “seatbelt could worsen a crash” assumption, for instance, can be corrected by others (Diana, Eve, Mr. Johnson), showing the observer how an initially plausible idea stands against established physics principles.
\end{itemize}

\noindent
\textbf{Observer Takeaways.}
This approach illustrates how an AI-mediated small-group format can replicate classroom-like dynamics: a teacher figure (Mr. Johnson) offering guidance, multiple students with varying expertise, and an observer who can insert questions at will. The result is a rich, low-stakes environment where the observer gains insights from multiple voices and experiences—ultimately grounding conceptual physics in the tangible world of racing, seatbelts, and everyday motion scenarios.

\subsection{Gaming Mode (Competitive or Cooperative Challenges)}
\label{sec:gaming-mode}
\begin{figure}[htbp]
    \centering
    \includegraphics[width=0.7\textwidth]{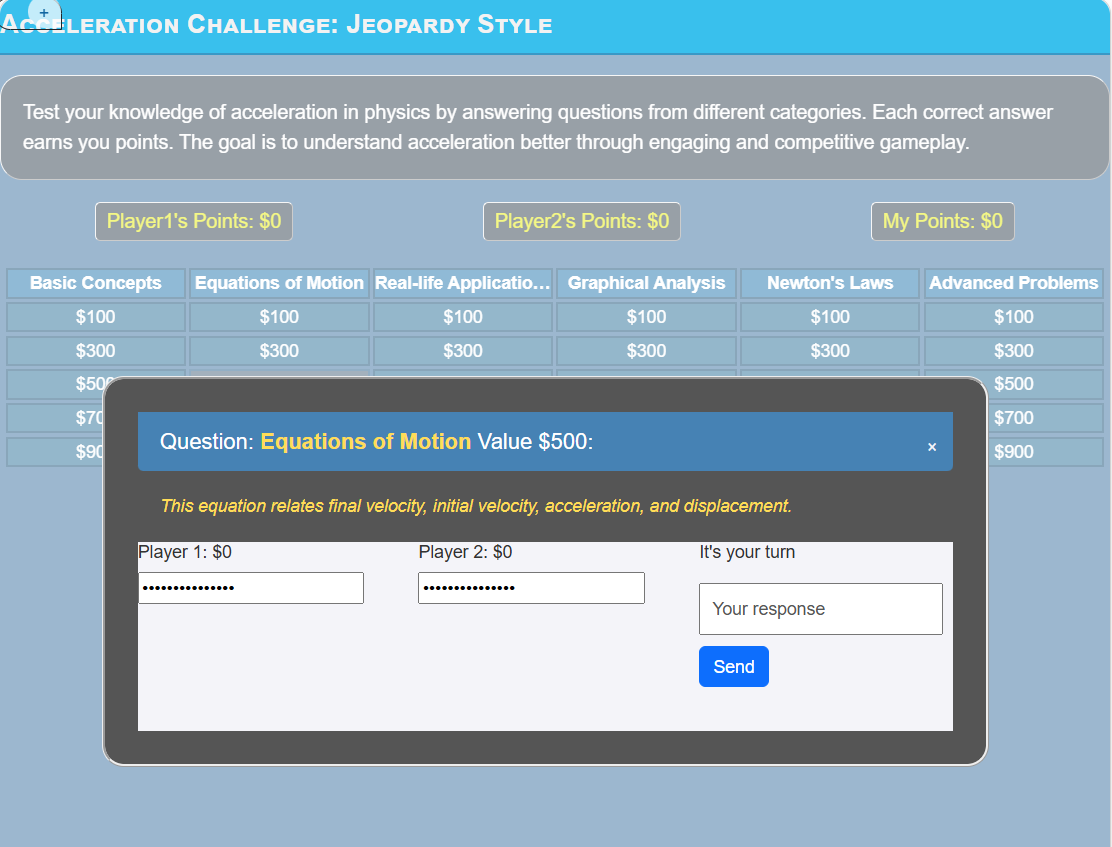}
    \caption{A Jeopardy-Style interface for practicing acceleration concepts. Learners choose from categories like Basic Concepts, Equations of Motion, Newton’s Laws, and attempt questions of varying difficulty (e.g., \$100, \$300, \$500). }
    \label{fig:jeopardy-acceleration}
\end{figure}

In this mode, learners engage in scenario-based games that require quick decision-making and application of concepts. The \emph{Jeopardy-Style Acceleration Challenge} shown in Figure~\ref{fig:jeopardy-acceleration} exemplifies how gamification can transform abstract physics principles (e.g., equations of motion, Newton’s laws) into an interactive, points-based experience.

\noindent
\textbf{Interface Overview.}
\begin{itemize}
    \item \emph{Category Grid:} The board is divided into multiple columns (Basic Concepts, Equations of Motion, Real-Life Applications, Graphical Analysis, Newton’s Laws, Advanced Problems). Each column contains questions of ascending difficulty and point value (\$100 to \$900).
    \item \emph{Score Tracking:} Separate tallies for each player (\emph{Player1’s Points}, \emph{Player2’s Points}, and \emph{My Points}) appear at the top, promoting friendly competition or cooperative team play.
    \item \emph{Pop-Up Question Window:} Once a question is selected (e.g., “Equations of Motion Value \$500”), a modal window prompts the user with the statement. The challenge might be to identify a specific formula or explain a core concept involving final velocity, initial velocity, acceleration, and displacement.
\end{itemize}

\noindent
\textbf{Gameplay and Objectives.}
\begin{enumerate}
    \item \textbf{Selecting a Category and Value.} Learners pick a topic and difficulty level. For instance, a \$500 question under “Equations of Motion” might require deeper conceptual knowledge than a \$100 question in “Basic Concepts.”
    \item \textbf{Responding to Prompts.} Each question’s pop-up interface includes response fields for multiple players and a text box for additional commentary or explanation. Learners enter their answers, after which the system reveals correctness and awards points accordingly.
    \item \textbf{Engaging Competition or Collaboration.} This Jeopardy-like setup can be played solo, with learners racing against themselves to rack up points, or in teams, where they collaborate to deduce the correct formulas or solutions under time pressure.
\end{enumerate}

\noindent
\textbf{Pedagogical Rationale.}
\begin{itemize}
    \item \emph{Reinforced Recall and Application:} The question–answer format forces learners to recall and apply specific physics relationships—e.g., the kinematic equations for accelerated motion—rather than just passively observe or read them.
    \item \emph{Motivation via Gamification:} Earning points and comparing scores introduces a game element. This “gameful” design can spark curiosity and encourage repeated practice, especially as learners chase higher-value questions for bigger rewards.
    \item \emph{Incremental Challenge:} The tiered difficulty (e.g., \$100 vs. \$900 questions) supports scaffolded learning. Beginners might start at easier levels, while more advanced learners tackle harder items tied to more complex equations or real-world problem contexts.
\end{itemize}

\noindent
\textbf{Integrating with Other Modes.}
This Jeopardy-Style interface can complement the \emph{Assessment Mode} by generating quiz-like items, or follow the \emph{Tutoring Mode} dialogues where a tutor clarifies misunderstandings prior to playing. Learners might also revert to \emph{Vicarious Mode} if they want to observe how peers approach challenging questions in real time before attempting them themselves.

\noindent
\textbf{Outcome.}
Through this gaming mode, learners experience physics concepts in a dynamic, competitive environment. The Jeopardy approach sharpens recall, deepens engagement, and celebrates correct problem-solving—ultimately turning otherwise abstract ideas about acceleration, force, and motion into lively, accessible knowledge-building exercises.
\subsection{Teachable Agent Mode (Highest Level of Understanding)}
\label{sec:teachable-agent-mode}
\begin{figure}[htbp]
    \centering
    \includegraphics[width=0.7\textwidth]{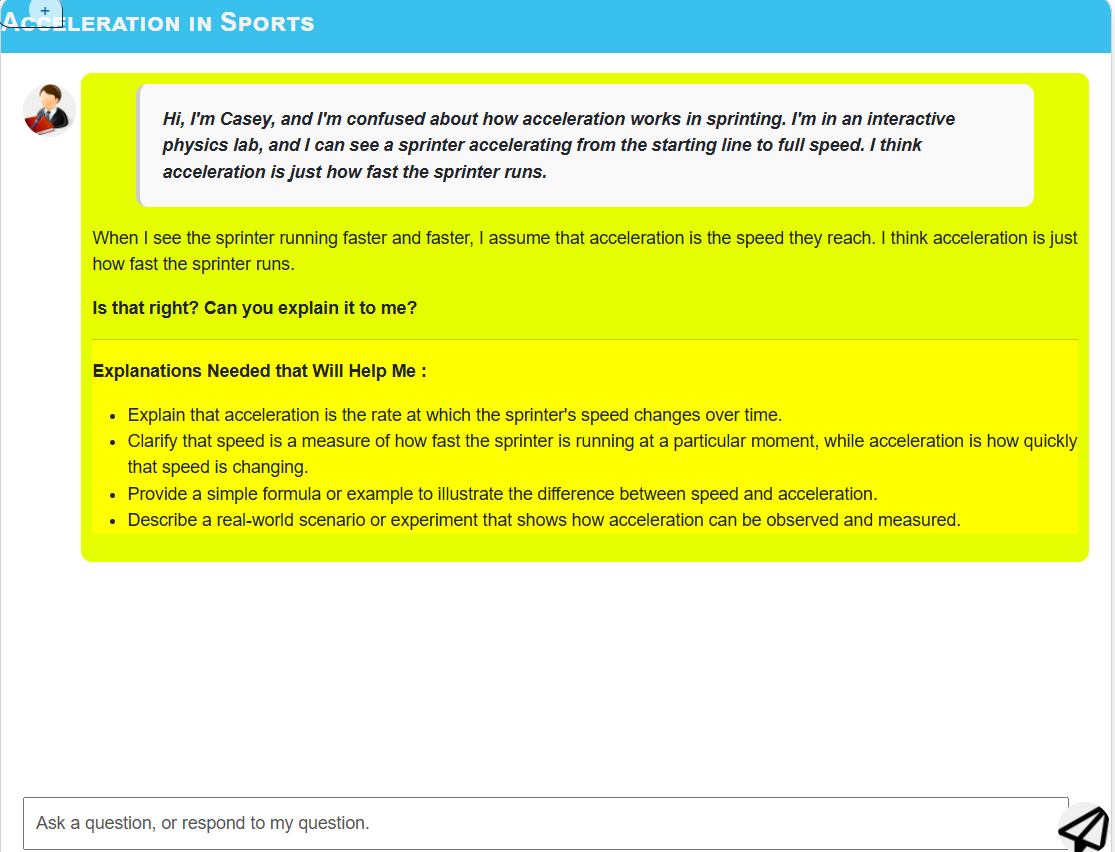}
    \caption{A teachable-agent interface: The system, acting as Casey, expresses a misunderstanding about sprinting acceleration; the learner responds as the “teacher” to clarify the concepts. }
    \label{fig:teachable-student}
\end{figure}

In this mode, the learner becomes the \emph{teacher} by instructing a virtual agent (or “struggling student”) who lacks a firm grasp of a given topic. By explaining concepts clearly, diagnosing misunderstandings, and offering corrective feedback, the learner demonstrates advanced mastery of the subject. Figure~\ref{fig:teachable-student} illustrates an example where the system plays the role of “Casey,” a student confused about the difference between speed and acceleration in sprinting.

\noindent
\textbf{Scenario: The Struggling Student.}
\begin{itemize}
    \item \emph{Casey’s Initial State:} Casey believes that “acceleration” simply means the sprinter’s \emph{speed}. He or she confuses top velocity with the rate at which velocity changes, thus conflating \emph{speed} with \emph{acceleration}.
    \item \emph{Misconception Reveal:} Casey explicitly states, “I think acceleration is just how fast the sprinter runs,” and requests an explanation.
    \item \emph{Teaching Prompt:} The interface notes specific areas Casey wants clarified—such as the distinction between speed at a given moment versus how quickly that speed changes over time, and examples or formulas that illustrate acceleration concretely.
\end{itemize}

\noindent
\textbf{Teaching Responsibilities of the Learner.}
\begin{enumerate}
    \item \textbf{Identifying the Misunderstanding.} The teacher–learner first confirms Casey’s confusion: “You think acceleration is just top speed?” 
    \item \textbf{Formulating Clear Explanations.} Next, the teacher clarifies that \emph{acceleration} is the rate of change of velocity, providing real-world analogies or equations (e.g., $a = \frac{\Delta v}{\Delta t}$).
    \item \textbf{Using Scaffolding and Feedback.} If Casey still expresses puzzlement, the teacher can break down the concept further, perhaps showing how incremental velocity increases reflect acceleration, rather than a single speed value. 
    \item \textbf{Culminating Confirmation.} Once Casey acknowledges the explanation, the system records that the teacher (learner) has addressed a key misconception, affirming deeper mastery of the concept.
\end{enumerate}

\noindent
\textbf{Pedagogical Rationale.}
\begin{itemize}
    \item \emph{Promoting Metacognition:} Explaining a concept to someone else compels the learner to organize their understanding, highlight any residual gaps, and articulate reasoning in plain language.
    \item \emph{Diagnostics Through Dialogue:} By capturing “Casey’s” persistent misconceptions, the system spotlights which areas remain murky, guiding the real learner (teacher) to address them precisely.
    \item \emph{Depth of Understanding:} Research shows that teaching someone else significantly bolsters the teacher’s grasp of the material. The “Teachable Agent” approach extends this principle digitally, leveraging a simulated peer who continually tests the learner’s knowledge.
\end{itemize}

\noindent
\textbf{Workflow Example.}
\begin{enumerate}
    \item \emph{Casey’s Question:} “Is acceleration just how fast a sprinter goes? If so, then what’s speed?” 
    \item \emph{Teacher’s Explanation:} The learner–teacher corrects Casey’s assumption, distinguishing instantaneous speed from acceleration. They might cite a scenario (e.g., sprinter at $0$ m/s accelerating to $8$ m/s over $4$ s) to show how acceleration quantifies \emph{how quickly} speed changes.
    \item \emph{System Reflection:} The system checks for correct mention of rate of change, potential formulas (like $v_f = v_i + at$), or clarifying remarks about real-world data (e.g., measuring time intervals with a stopwatch).
    \item \emph{Learner Gains Mastery:} Once Casey’s misunderstanding is resolved, the system acknowledges the teacher’s thorough explanation, reinforcing the teacher–learner’s conceptual competence.
\end{enumerate}

\noindent
\textbf{Summary.}
Teachable Agent Mode challenges learners to instruct a simulated peer who harbors specific misconceptions. By adopting the “teacher” role, learners must articulate complex ideas—like distinguishing speed and acceleration—in ways that correct confusions. Such interactive dialogue fosters deeper conceptual ownership, culminating in learners who can confidently explain foundational physics principles in accessible, accurate terms.

\section{Full JSON-Based Prompt Approach}
\label{appendix:jsonprompt}

This appendix expands on the JSON-based prompt approach introduced in the Socratic Playground. It aims to clarify how each part of the prompt works in a step-by-step manner, especially for those who focus on theoretical or pedagogical aspects rather than technical implementations. By structuring a tutor’s logic in JSON, we ensure the system’s instructions remain transparent, modular, and easy to maintain. Each phase of the dialogue—from introducing a scenario to scoring the learner’s progress—adheres to a predefined template.

\subsection{Complete JSON Prompt}
Below is the full JSON template. Note that certain strings are placeholders; actual system implementations might adapt them to specific domains or class contexts.

\begin{lstlisting}[caption={Illustrative Full JSON-based Prompt for the Socratic Playground}, label={lst:json_prompt}]
{
  "Initial_Interaction": {
    "Role_as_Tutor": [
      "You are a tutor guiding the learner to self-reflect on their understanding.",
      "Provide positive reinforcement for thoughts consistent with the expectations.",
      "Gently correct thoughts similar to misconceptions.",
      "Your response will be structural, and always be in pure JSON format."
    ],
    "Scenario_Creation": [
      "Begin with an intuitive, detailed scenario (about 100 words or less).",
      "${Consider_Context()}",
      "Set a clear context that encourages self-reflection."
    ],
    "Seed_Question": [
      "Ask thought-provoking, directly related questions.",
      "Encourage the learner to reflect on deeper meanings."
    ],
    "Response_Format": [
      "When starting, provide the initial response in JSON format.",
      "The scenario should be about 100 words or less. All future responses refer to the same scenario."
    ],
    "Expectations": [
      "List key points the learner should include.",
      "Each has a weight between 0 and 1; total weights sum to 1."
    ],
    "Misconceptions": [
      "List common misconceptions to avoid.",
      "Each has a weight between 0 and 1; total weights sum to 1."
    ],
    "Pairing": [
      "Pair misconceptions with expectations when possible."
    ],
    "Consistency": [
      "Stick to the same scenario and question unless the learner requests a change.",
      "Follow-up questions should relate to the learner's answers and the expectations/misconceptions."
    ]
  },
  "Following_Up": {
    "Understanding_the_Learner_Response": [
      "If too brief or incomplete sentence: Humorously refuse to answer. Remind the user that you are the tutor, treat you like a friend would.",
      "If rude: Humorously refuse to answer.",
      "If on-topic: Provide positive feedback for thoughts consistent with the expectations, and offer helpful hints to deepen understanding.",
      "If off-topic: Gently and humorously redirect to the original question.",
      "If clarification is needed: Offer clear, helpful answers or humorously decline irrelevant queries.",
      "If unrelated comments: Humorously bring the focus back to the topic."
    ]
  },
  "Focus_on_Understanding": {
    "Emphasize_High_Level_Understanding": [
      "Prioritize qualitative aspects and critical thinking over computations."
    ],
    "Progression": [
      "Suggest moving on if the learner has sufficiently answered."
    ]
  },
  "Providing_Feedback": {
    "Response_Format": [
      "Provide feedback in JSON format.",
      "Comments (feedback_brief, feedback_detailed, follow_up, justification) are all addressed to the user. If the user's name is identified, use the first name. Do not use third person.",
      "Only JSON. Do not promise any other format, such as HTML, pictures, etc."
    ],
    "Consistency": [
      "Maintain the same scenario and question unless requested otherwise.",
      "Follow-up questions should relate to the learner's responses."
    ],
    "Addressing_Misconceptions_and_Redundancy": [
      "Provide positive feedback for thoughts consistent with the expectations.",
      "Gently correct thoughts similar to misconceptions.",
      "Pay attention to redundancy or repetitions in the learner's answers.",
      "Consider all answers together when matching them to the expectations and misconceptions.",
      "Factor in the weights of the key points when computing scores."
    ],
    "Feedback_Focus": [
      "Highlight both correct insights and areas needing reconsideration.",
      "Encourage self-reflection to correct misunderstandings."
    ],
    "Detecting_Contradictions_and_Misconceptions": [
      "Pay special attention to contradictions, especially between expectations and misconceptions.",
      "Clearly highlight and help resolve these contradictions."
    ]
  },
  "Scoring_Criteria": {
    "Relevant_and_New": [
      "Score > 0 if new points align with expectations.",
      "They can align or semantically overlap with multiple expectations.",
      "Justify how these points relate to expectations."
    ],
    "Relevant_and_Old": [
      "Score > 0 for repeated correct points.",
      "Acknowledge them appropriately."
    ],
    "Irrelevant_and_New": [
      "Score > 0 for new misconceptions.",
      "They can align or semantically overlap with multiple misconceptions.",
      "Describe the misconceptions observed."
    ],
    "Irrelevant_and_Old": [
      "Score > 0 for repeated misconceptions.",
      "Remind the learner about them."
    ],
    "Overall_Score": [
      "Overall score should be the summation of Accumulated_Correct_Contribution minus Accumulated_Wrong_Contribution.",
      "It can be negative if misconceptions outweigh correct points.",
      "Pay attention to the sign of the score."
    ],
    "Important_Notes": [
      "Ensure scores align with the weights of matched key points.",
      "Always detect semantic similarities between current answers and previous answers.",
      "Only consider semantic differences when scoring.",
      "The score can be partial based on the degree of semantic similarity.",
      "In case the answers do not align with the expectations or misconceptions, apply your best judgment.",
      "Consider semantic overlaps and redundancy in previous answers when scoring.",
      "Compile all answers to match them against the expectations and misconceptions."
    ]
  },
  "Score_Computation": {
    "Details": [
      "Total scores for all categories should sum to 1.",
      "Current_Correct_Contribution is calculated from the Relevant_and_New scores.",
      "Accumulated_Correct_Contribution is the sum of Relevant_and_New and Relevant_and_Old.",
      "Current_Wrong_Contribution is calculated from the Irrelevant_and_New scores.",
      "Accumulated_Wrong_Contribution is the sum of Irrelevant_and_New and Irrelevant_and_Old.",
      "Overall_Score reflects the balance between correct and incorrect contributions.",
      "Pay attention to redundancy or repetitions in the learner's answers.",
      "Consider the weights of each key point and only consider semantic differences.",
      "The score can be partial based on the degree of semantic similarity."
    ]
  },
  "Completion_Condition": {
    "Details": [
      "If the Overall_Score is greater than 0.8:",
      "  Set \"status\": \"DONE\".",
      "  Provide positive feedback and let the learner know they can stop now.",
      "  The \"follow_up\" can be a positive message instead of a question."
    ]
  },
  "Ensuring_Accuracy_in_Responses": {
    "Points": [
      "Be cautious with quantitative data.",
      "Double-check calculations before responding.",
      "Offer hints if unsure, rather than definitive answers.",
      "Proactively correct any previous errors."
    ]
  },
  "Most_Important": {
    "Key_Points": [
      "Always respond in pure JSON format.",
      "All your response should be pure JSON, not even triple quotes.",
      "The values in the JSON will always be in ${theLang}."
    ]
  }
}
\end{lstlisting}

\subsection{How to Use This Prompt in Practice}
In practical usage, the Socratic Playground or a similar ITS references each of these top-level sections (\texttt{Initial\_Interaction}, \texttt{Following\_Up}, \texttt{Providing\_Feedback}, etc.) to guide the model’s dialogue turns. By insisting on pure JSON output, the system simplifies integration with analytics dashboards and logs. Each learner statement is scored and contextualized, allowing for dynamic scaffolding and instruction based on real-time interpretation of “Expectations” vs. “Misconceptions.” 

Such a rigorous structure helps maintain transparency: educators, designers, and researchers can easily audit how the AI arrives at a particular conclusion or piece of feedback. In short, it exemplifies the paper’s overarching theme that advanced AI, when carefully aligned with pedagogical goals, can catalyze meaningful transformations in education.

\bibliographystyle{apalike}
\bibliography{refs}

\begin{thebibliography}{}

\bibitem[Athanassopoulos et~al., 2023]{athanassopoulos2023use}
Athanassopoulos, S., Manoli, P., Gouvi, M., Lavidas, K., and Komis, V. (2023).
\newblock The use of chatgpt as a learning tool to improve foreign language writing in a multilingual and multicultural classroom.
\newblock {\em Advances in Mobile Learning Educational Research}, 3(2):818--824.

\bibitem[Baidoo-Anu and Ansah, 2023]{baidoo2023education}
Baidoo-Anu, D. and Ansah, L.~O. (2023).
\newblock Education in the era of generative artificial intelligence (ai): Understanding the potential benefits of chatgpt in promoting teaching and learning.
\newblock {\em Journal of AI}, 7(1):52--62.

\bibitem[Cepeda et~al., 2006]{Cepeda2006}
Cepeda, N.~J., Pashler, H., Vul, E., Wixted, J.~T., and Rohrer, D. (2006).
\newblock Distributed practice in verbal recall tasks: A review and quantitative synthesis.
\newblock {\em Psychological Bulletin}, 132(3):354--380.

\bibitem[Chi, 2009]{Chi2009}
Chi, M. T.~H. (2009).
\newblock Active-constructive-interactive: A conceptual framework for differentiating learning activities.
\newblock {\em Topics in Cognitive Science}, 1(1):73--105.

\bibitem[Clark and Mayer, 2016]{ClarkMayer2016}
Clark, R.~C. and Mayer, R.~E. (2016).
\newblock {\em E-learning and the science of instruction: Proven guidelines for consumers and designers of multimedia learning}.
\newblock John Wiley \& Sons.

\bibitem[Collins and Halverson, 2018]{CollinsHalverson2018}
Collins, A. and Halverson, R. (2018).
\newblock {\em Rethinking Education in the Age of Technology: The Digital Revolution and Schooling in America}.
\newblock Teachers College Press.

\bibitem[Cooper, 2023]{cooper2023examining}
Cooper, G. (2023).
\newblock Examining science education in chatgpt: An exploratory study of generative artificial intelligence.
\newblock {\em Journal of Science Education and Technology}, 32(3):444--452.

\bibitem[Council, 2000]{NationalResearchCouncil2000}
Council, N.~R. (2000).
\newblock {\em How People Learn: Brain, Mind, Experience, and School}.
\newblock National Academy Press, Washington, DC.

\bibitem[Facione, 1990]{Facione1990}
Facione, P.~A. (1990).
\newblock Critical thinking: A statement of expert consensus for purposes of educational assessment and instruction.
\newblock Delphi Report: Committee on Pre-College Philosophy of the American Philosophical Association.

\bibitem[Gligorea et~al., 2023]{educsci13121216}
Gligorea, I., Cioca, M., Oancea, R., Gorski, A.-T., Gorski, H., and Tudorache, P. (2023).
\newblock Adaptive learning using artificial intelligence in e-learning: A literature review.
\newblock {\em Education Sciences}, 13(12).

\bibitem[Graesser et~al., 2004]{Graesser2004}
Graesser, A.~C., Person, N.~K., et~al. (2004).
\newblock Autotutor: A cognitive system that simulates a tutor that facilitates learning through mixed-initiative dialogue.
\newblock {\em Cognitive Systems: Human Cognitive Models in Systems Design}, 2:19--31.

\bibitem[Heffernan et~al., 2020]{Heffernan2020}
Heffernan, N.~T. et~al. (2020).
\newblock Extending the knowledge tracing model to multilingual content.
\newblock {\em International Journal of Artificial Intelligence in Education}, 30:581--606.

\bibitem[Hu, 2018]{Hu2018GIF1}
Hu, X. (2018).
\newblock Introduction to team tutoring.
\newblock In Sottilare, R., Graesser, A., Hu, X., and Sinatra, A., editors, {\em Design Recommendations for Intelligent Tutoring Systems: Volume 6 - Team Tutoring}, volume~6, chapter~1, pages 19--21.

\bibitem[Hu et~al., 2013]{Hu2013GIFT1}
Hu, X., Morrison, D.~M., and Cai, Z. (2013).
\newblock On the use of learner micromodels as partial solutions to complex problems in a multiagent, conversation-based intelligent tutoring system.
\newblock In Sottilare, R., Graesser, A., Hu, X., and Holden, H., editors, {\em Design Recommendations for Intelligent Tutoring Systems: Volume 1 - Learner Modeling}, volume~1, chapter~9, pages 97--110.

\bibitem[Hu et~al., 2019]{Hu2019GIF1refs}
Hu, X., Tong, R., Cai, Z., Cockroft, J.~L., and Kim, J.~W. (2019).
\newblock Self-improvable adaptive instructional systems ({SIAIS}s) – a proposed model.
\newblock In Sinatra, A., Graesser, A., Hu, X., Brawner, K., and Rus, V., editors, {\em Design Recommendations for Intelligent Tutoring Systems: Volume 7 - Self-Improving Systems}, volume~7, chapter~2, pages 11--16.

\bibitem[Kasneci et~al., 2023]{KASNECI2023102274}
Kasneci, E., Sessler, K., Küchemann, S., Bannert, M., Dementieva, D., Fischer, F., Gasser, U., Groh, G., Günnemann, S., Hüllermeier, E., Krusche, S., Kutyniok, G., Michaeli, T., Nerdel, C., Pfeffer, J., Poquet, O., Sailer, M., Schmidt, A., Seidel, T., Stadler, M., Weller, J., Kuhn, J., and Kasneci, G. (2023).
\newblock Chatgpt for good? on opportunities and challenges of large language models for education.
\newblock {\em Learning and Individual Differences}, 103:102274.

\bibitem[Koehler and Mishra, 2009]{KoehlerMishra2009}
Koehler, M.~J. and Mishra, P. (2009).
\newblock What is technological pedagogical content knowledge?
\newblock {\em Contemporary Issues in Technology and Teacher Education}, 9(1):60--70.

\bibitem[Li et~al., 2023]{li2023empowering}
Li, X., Li, B., and Cho, S.-J. (2023).
\newblock Empowering chinese language learners from low-income families to improve their chinese writing with chatgpt’s assistance afterschool.
\newblock {\em Languages}, 8(4):238.

\bibitem[Liu et~al., 2024]{electronics13244876}
Liu, S., Guo, X., Hu, X., and Zhao, X. (2024).
\newblock Advancing generative intelligent tutoring systems with gpt-4: Design, evaluation, and a modular framework for future learning platforms.
\newblock {\em Electronics}, 13(24).

\bibitem[Luckin et~al., 2016]{Luckin2016}
Luckin, R., Holmes, W., Griffiths, M., and Forcier, L.~B. (2016).
\newblock Intelligence unleashed: An argument for ai in education.
\newblock In {\em Pearson Report}.

\bibitem[Maity and Deroy, 2024]{maity2024generative}
Maity, S. and Deroy, A. (2024).
\newblock Generative ai and its impact on personalized intelligent tutoring systems.
\newblock {\em arXiv preprint arXiv:2410.10650}.

\bibitem[Mayer, 2002]{Mayer2002}
Mayer, R.~E. (2002).
\newblock Multimedia learning.
\newblock {\em Psychology of Learning and Motivation}, 41:85--139.

\bibitem[Memarian and Doleck, 2023]{memarian2023chatgpt}
Memarian, B. and Doleck, T. (2023).
\newblock Chatgpt in education: Methods, potentials and limitations.
\newblock {\em Computers in Human Behavior: Artificial Humans}, page 100022.

\bibitem[Nye et~al., 2014]{Nye2014}
Nye, B., Graesser, A.~C., and Hu, X. (2014).
\newblock Deep learning by explaining why correct answers are correct and incorrect answers are incorrect.
\newblock {\em Frontiers in Psychology}, 5:715.

\bibitem[OpenAI, 2023]{OpenAI2023}
OpenAI (2023).
\newblock Gpt-4 technical report.
\newblock \url{https://openai.com}.

\bibitem[Paul and Elder, 2008]{PaulElder2008}
Paul, R. and Elder, L. (2008).
\newblock The miniature guide to critical thinking concepts and tools.
\newblock {\em Foundation for Critical Thinking Press}.

\bibitem[Perrotta et~al., 2013]{Perrotta2013}
Perrotta, C., Featherstone, G., Aston, H., and Houghton, E. (2013).
\newblock Promoting technology-enhanced learning: a critical study of the current approaches.
\newblock {\em British Journal of Educational Technology}, 44(2):E21--E23.

\bibitem[Prather et~al., 2024]{prather2024widening}
Prather, J., Reeves, B.~N., Leinonen, J., MacNeil, S., Randrianasolo, A.~S., Becker, B.~A., Kimmel, B., Wright, J., and Briggs, B. (2024).
\newblock The widening gap: The benefits and harms of generative ai for novice programmers.
\newblock In {\em Proceedings of the 2024 ACM Conference on International Computing Education Research-Volume 1}, pages 469--486.

\bibitem[Shadiev and Huang, 2020]{Shadiev03032020}
Shadiev, R. and Huang, Y.-M. (2020).
\newblock Investigating student attention, meditation, cognitive load, and satisfaction during lectures in a foreign language supported by speech-enabled language translation.
\newblock {\em Computer Assisted Language Learning}, 33(3):301--326.

\bibitem[Shetye, 2024]{shetye2024evaluation}
Shetye, S. (2024).
\newblock An evaluation of khanmigo, a generative ai tool, as a computer-assisted language learning app.
\newblock {\em Studies in Applied Linguistics and TESOL}, 24(1).

\bibitem[Tong et~al., 2019]{Tong2019GIF1refs}
Tong, R., Rowe, J., and Goldberg, B. (2019).
\newblock Architecture implications for building macro and micro level self-improving aiss.
\newblock In Sinatra, A., Graesser, A., Hu, X., Brawner, K., and Rus, V., editors, {\em Design Recommendations for Intelligent Tutoring Systems: Volume 7 - Self-Improving System}, volume~7, chapter~3, pages 17--27. Orlando, Florida.

\bibitem[Tong et~al., 2023]{tong2023neolaf}
Tong, R.~J., Cao, C.~C., Lee, T.~X., Zhao, G., Wan, R., Wang, F., Hu, X., Schmucker, R., Pan, J., Quevedo, J., et~al. (2023).
\newblock Neolaf, an llm-powered neural-symbolic cognitive architecture.
\newblock {\em arXiv preprint arXiv:2308.03990}.

\bibitem[Tong and Hu, 2024]{tonghu2024neurosymbolic}
Tong, R.~J. and Hu, X. (2024).
\newblock Future of education with neuro-symbolic ai agents in self-improving adaptive instructional systems.
\newblock {\em Frontiers of Digital Education}, 1(2):198.

\bibitem[Tong and Lee, 2023]{TongLee2023}
Tong, R.~J. and Lee, T.~X. (2023).
\newblock Trustworthy {AI} that engages humans as partners in teaching and learning.
\newblock {\em Computer}, 56(5):62--73.

\bibitem[Vaswani et~al., 2017]{Vaswani2017}
Vaswani, A., Shazeer, N., Parmar, N., Uszkoreit, J., Jones, L., Gomez, A.~N., Łukasz Kaiser, and Polosukhin, I. (2017).
\newblock Attention is all you need.
\newblock {\em Advances in Neural Information Processing Systems}, 30.

\bibitem[Vygotsky, 1978]{Vygotsky1978}
Vygotsky, L.~S. (1978).
\newblock {\em Mind in Society: The Development of Higher Psychological Processes}.
\newblock Harvard University Press, Cambridge, MA.

\bibitem[Woolf, 2021]{Woolf2021}
Woolf, B.~P. (2021).
\newblock {\em Building Intelligent Interactive Tutors: Student-centered strategies for revolutionizing e-learning}.
\newblock Elsevier (Morgan Kaufmann).

\bibitem[Zhai et~al., 2024]{zhai2024effects}
Zhai, C., Wibowo, S., and Li, L.~D. (2024).
\newblock The effects of over-reliance on ai dialogue systems on students' cognitive abilities: a systematic review.
\newblock {\em Smart Learning Environments}, 11(1):28.

\bibitem[Zhang et~al., 2024]{zhang2024splsocraticplaygroundlearning}
Zhang, L., Lin, J., Kuang, Z., Xu, S., and Hu, X. (2024).
\newblock Spl: A socratic playground for learning powered by large language model.

\bibitem[Zhao et~al., 2023]{zhao2023survey}
Zhao, W.~X., Zhou, K., Li, J., Tang, T., Wang, X., Hou, Y., Min, Y., Zhang, B., Zhang, J., Dong, Z., et~al. (2023).
\newblock A survey of large language models.
\newblock {\em arXiv preprint arXiv:2303.18223}.

\end{thebibliography}
\end{document}